\pdfoutput=1

\documentclass[11pt]{article}

\usepackage[]{emnlp2021}

\usepackage{times}
\usepackage{latexsym}
\usepackage{graphicx}
\usepackage{booktabs}
\usepackage{xcolor}
\usepackage{caption}
\usepackage{subcaption}
\usepackage{multirow}
\usepackage[T1]{fontenc}

\usepackage[utf8]{inputenc}

\usepackage{microtype}

\newcommand{\methodname}{\textsc{REDE}}

\title{Towards Zero and Few-shot Knowledge-seeking Turn Detection in Task-orientated Dialogue Systems}

\author{Di Jin, Shuyang Gao, Seokhwan Kim, Yang Liu, Dilek Hakkani-Tur \\
        Amazon Alexa AI, Sunnyvale, CA, USA\\
        \texttt{\{djinamzn,shuyag,seokhwk,yangliud,hakkanit\}@amazon.com}}

\begin{document}
\maketitle
\begin{abstract}
Most prior work on task-oriented dialogue systems is restricted to  supporting domain APIs. However, users may have requests that are out of the scope of these APIs. This work focuses on identifying such user requests. Existing methods for this task mainly rely on fine-tuning pre-trained models on large annotated data. We propose a novel method, {\methodname}, based on adaptive representation learning and density estimation. {\methodname} can be applied to zero-shot cases, and quickly learns a high-performing detector with only a few shots by updating less than 3K parameters. We demonstrate {\methodname}'s competitive performance on DSTC9 data and our newly collected test set. 

\end{abstract}

\section{Introduction}

Current task-oriented dialog systems often rely on pre-defined APIs to complete target tasks~\cite{williams:2017,eric-manning:2017} and filter out any other requests beyond the APIs as out-of-domain cases.
However, some in-domain user requests can be addressed by incorporating external domain knowledge from the web or any other sources~\cite{kim-etal-2020-beyond}. 
To address this problem, \citet{kim2021beyond} recently organized a benchmark challenge on task-oriented conversational modeling with unstructured knowledge access in DSTC9~\cite{DBLP:journals/corr/abs-2011-06486}. 
This challenge includes the knowledge-seeking turn detection task to determine whether to invoke a knowledge-driven responder or just rely on available API functions. One data sample is provided in Table \ref{table:data-sample}.
The state-of-the-art systems~\cite{he2021learning,tang2021radge,mi2021towards,jin-etal-2021-assistance} implemented this detector by fine-tuning a large pre-trained model 
on the training dataset (about 72K samples) as a binary classifier, and achieved an F1 score of over 95\% on the benchmark test set. However, after close investigation, we find  those user queries in the test set are very limited in topic coverage and language variation.  To evaluate the detector performance on real-world user queries, we specially curate a new contrast set following~\citet{gardner-etal-2020-evaluating} by manually collecting questions posted by real users on Tripadvisor forums. 
We found that the detector trained on DSTC9 Track 1 training samples had a large performance degradation on this contrast set (F1 score dropped by over 15\%), suggesting the need for methods with better generalization. 

\begin{table*}[t]
\small
\centering
\resizebox{0.99\textwidth}{!}{
\begin{tabular}{p{0.02\textwidth}p{0.05\textwidth}p{0.51\textwidth}p{0.41\textwidth}}
\toprule
Turn & Speaker & Utterance & Sampled Knowledge Snippets from FAQs \\ \midrule
1     & User        & I'm looking for a place to stay in the south of town. It doesn't need to have free parking.          &                            \\
2     & Agent        &  There are 4 hotels that are in the area you are looking for. Would you prefer a 3 or 4 star rated hotel?         &                            \\
3     & User        &   I don't care about the star rating as long as it's expensive.        &                            \\
4     & Agent        &  The \textbf{\textit{Lensfield Hotel}} is the only expensive hotel in the south area. Would you like any more information on this location?         &                            \\
5     & User        &  I'm interested in knowing, do they have a workout facility on the premises?         &                            \\
6     & Agent        &  There are both a fitness center and gym available on the premises. Does this sound ok?         & Q1: Do you have room service for your guests?\\
& & & A1: Yes, the Lensfield Hotel provides room services.\\
& & & \textbf{Q2: Is there a gym available at your location?} \\
& & & \textbf{A2: There is both a fitness center and gym available on the premises.} \\
& & & Q3: Can I bring my dog? \\
& & & A3: Pets are not allowed at the Lensfield Hotel. \\
7     & User        &  That is perfect can you book that for me please.         &                            \\
8     & Agent        &   The Lensfield Hotel is located in the South. It has a 3 star rating and is expensive. There is free parking and internet. I have booked it for you.        &                            \\
9     & User        & Great, thank you!          &                           \\ \bottomrule
\end{tabular}}
\caption{One example of task-oriented conversations with unstructured knowledge access. The user utterance of turn 5 is beyond the coverage of API and needs external knowledge support, therefore it is a knowledge-seeking turn while other user turns are non-knowledge-seeking turns. To address turn 5, three sampled FAQ pairs for the entity ``Lensfield Hotel'' extracted from the external knowledge are listed in the rightmost column. The most appropriate FAQ pair is highlighted in bold font.}
\label{table:data-sample}
\end{table*}

In this work, we propose a method that can quickly learn a knowledge-seeking turn detector with much fewer out-of-domain samples, such as only a few shots or even zero shot.  
Our method is composed of two stages: \textbf{RE}presentation learning and \textbf{DE}nsity estimation ({\methodname}). 
First, we learn a representation model via fine-tuning a pre-trained sentence encoder 
on all non-knowledge-seeking turns (utterances that can be supported by APIs) via masked language modeling (MLM). 
Then we learn a density estimator using these representation vectors. 
During inference, the density estimator produces a density score for a given user utterance. If it is above a threshold, this utterance is counted as an in-domain API turn, otherwise as a knowledge-seeking turn. 

To incorporate out-of-domain examples, we propose to use principle component analysis to quickly learn a projection matrix with few knowledge-seeking turn samples and then use this matrix to linearly transform the representation vectors.
We conduct experiments on the DSTC9 Track 1 data as well as our new contrast test set. We demonstrate that {\methodname} can achieve competitive performance as other supervised methods in the full-shot setting and outperform them by a large margin in the low-resource setting.
More importantly, our approach generalizes much better in the new contrast test set that we created. 

Overall, our contributions are summarized as follows:

\begin{itemize}
    \item We propose a new approach, {\methodname}, for knowledge-seeking turn detection that be applied to zero or few shot cases.  
    It can be quickly adapted to new knowledge-seeking turns data with much less training samples,
    which can achieve over 90\% F1 score with only five shots;
    \item Once the continuous pretraining stage on non-knowledge-seeking turns data is finished, our model can be quickly adapted to any kinds of knowledge-seeking turns data within seconds with only a few parameters to be learned.
    \item We curate and release a contrast set to examine the generalization capability of the knowledge seeking-turn detectors.\footnote{https://github.com/jind11/REDE} 
    We demonstrate that our model is better at generalizing to this contrast set than the previous best models. 
\end{itemize}

\section{Related Work}

Our work is closely related to those participating systems in DSTC9 Track 1~\cite{kim-etal-2020-beyond, kim2021beyond}. 
All the systems proposed to treat the problem of knowledge-seeking turn detection as a binary classification task and fine-tuned pre-trained models such as RoBERTa, UniLM, PLATO, GPT2, on the whole training set~\citep{he2021learning,tang2021radge,mi2021towards},  which yielded around 99\% and 96\% F1 scores on the development and test sets, respectively. Our method differs in two aspects: 1) We do not need to fine-tune the pre-trained model on the training set; 2) Our model is at least 5 times smaller and we need less than 5\% of training data to achieve similar performance.

Our method is inspired by previous work for out-of-domain (OOD) detection
\citep{NEURIPS2019_1e795968,gangal2020likelihood,hendrycks2018deep} and one-class classification~\citep{sohn2021learning}. \citet{kim-etal-2020-beyond} also tried tackling this problem by applying an unsupervised anomaly detection algorithm, Local Outlier Factor (LOF)~\citep{breunig2000lof}, which compares the local densities between a given input instance and its nearest neighbors, but did not obtain good results (F1 score is less than 50\%). \citet{sohn2021learning} proposed to first learn a representation model via contrastive learning, then learn a density estimator on the obtained representations. They showed decent performance for one-class classification. 
All these previous work assumed no access to OOD samples, however, we would like to make use of those OOD samples efficiently when they are available. Therefore we extend the general representation learning framework by proposing a novel representation transformation method to learn OOD samples, which leads to significantly boosted detection performance. 

Our work is also related to few/zero-shot learning, which has been widely studied previously~\citep{gao2020machine,jin2020simple,jin-etal-2020-hooks}. Transfer learning~\citep{zhou-etal-2019-dual,jin2020mmm,zhou2019dual,zhou2019roseq,yan-etal-2020-multi-source} and data augmentation~\citep{jindal2020speechmix,jindal-etal-2020-augmenting} have been two major methods for this direction, while our work focuses on manipulation of learned representations, which provides a new perspective.  

\section{Methods}

Our method includes three steps: encoder adaptation, representation transformation, and density estimation. 
The representation transformation step is only applicable when there are OOD examples (i.e., knowledge-seeking turns).

\subsection{Encoder Adaptation}

In this step, we adapt a pre-trained sentence encoder $E$ to the in-domain data, i.e., non-knowledge-seeking turns, $X^{NK}=\{x^{NK}_1,...,x^{NK}_N\}$, via masked language modeling~\citep{devlin-etal-2019-bert}. Specifically, 15\% of tokens of $x^{NK}_i$ are masked and $E$ is  trained to predict these masked tokens. 

\subsection{Representation Transformation}
\label{sec:rt}

To incorporate the knowledge-seeking turns $X^{K}=\{x^{K}_1,...,x^{K}_M\}$, a standard solution is to fine-tune $E$ on the combined data of knowledge-seeking and non-knowledge-seeking turns, $X=X^{K}\cup X^{NK}$, as a supervised binary classifier. However, in few-shot settings where $M << N$, there is an extreme class imbalance problem. In addition,  fine-tuning large models may take a long time and much computation power with large data size. Instead, we propose a simple linear transformation to the sentence representation $e=E(x)$ without updating the model parameters, following~\citep{su2021whitening}:
\begin{equation}
\label{eq:transformation}
    \tilde{e}=T(e)=(e-\mu)W
\end{equation}
where $\mu=\frac{1}{M}\sum_{i=1}^{M}E(x_i^{K})$.  
To calculate $W$, we first calculate the covariance matrix, $\Sigma=\frac{1}{M}\sum_{i=1}^{M}(E(x_i^{K})-\mu)^T(E(x_i^{K})-\mu)$, then perform Singular Value Decomposition (SVD) over $\Sigma$ such that: $\Sigma=U\Lambda U^T$, and finally we obtain $W=U\sqrt{\Lambda^{-1}}$. 



The elements in diagonal matrix $\Lambda$ derived from SVD are sorted in descending order.  Therefore, we can retain the first $L$ columns of $W$ to reduce the dimension of transformed vectors $\tilde{e}$, which is theoretically equivalent to Principal Component Analysis (PCA). However, to be noted, both $\mu$ and $W$ parameters are obtained using those knowledge-seeking turns instead of non-knowledge-seeking turns and the number of knowledge-seeking turns is much smaller, which can be as small as just a few shots. In another word, we only need a very small size of out-of-domain samples to learn the parameters needed for our representation transformation as defined in Eq. \ref{eq:transformation} to transform the representations of in-domain data. This is in contrast to the conventional PCA based density estimation method that assumes only having access to in-domain data, i.e. non-knowledge-seeking turns, and needs to learn and perform PCA transformation both on a good amount of those in-domain data. 

This step of transformation can be viewed as another round of  unsupervised representation learning with knowledge-seeking turns, which helps us obtain a better representation and is extremely critical to our claimed great performance for few-shot learning, as analyzed in Section~ \ref{sec:PCA-transformation}.

\subsection{Density estimation}
In this step, we encode all the non-knowledge-seeking turns in the training set and transform them to obtain $\{\tilde{e}^{NK}_1,...,\tilde{e}^{NK}_N\}$, normalize them into unit vectors, and then learn a shallow density estimator $D$ over them, such as Gaussian Mixture Model (GMM). 
Note that in the zero-shot setting when no knowledge-seeking turns are available, the representation transformation step (in Section \ref{sec:rt}) is skipped. 

During inference, given a test sample $x$, we encode it with the encoder $E$, transform it with $T$ defined in Eq. \ref{eq:transformation}, and then use the learned density estimator $D$ to produce a density score $D(T(E(x)))$. If it is above a pre-set threshold $\eta$, $x$ is considered as a non-knowledge-seeking turn, otherwise as a knowledge-seeking turn. This whole pipeline is motivated by the assumption that the well learned representations of in-domain (non-knowledge-seeking turns) and OOD samples (knowledge-seeking turns) should be distributed separately in the latent space, and thus the estimated density of in-domain data by the density estimator should be higher than that of OOD data.

\section{Experiments}

\subsection{Dataset}

We use the DSTC9 Track 1 competition data ~\citep{kim-etal-2020-beyond,kim2021beyond}, and focus on the sub-task of binary knowledge-seeking turn detection. The data statistics are summarized in Table \ref{tab:data-stats}.\footnote{Data can be downloaded from: https://github.com/alexa/alexa-with-dstc9-track1-dataset} We further curate a new contrast test set by first collecting questions posted by real users in the Tripadvisor forums\footnote{https://www.tripadvisor.com}, then obtain the questions as knowledge-seeking turns that cannot be addressed by MultiWOZ API schema~\citep{eric2019multiwoz} (this schema was also used for constructing the DSTC9 Track 1 dataset), and finally manually paraphrasing them to make them more like dialogue utterances. We  obtained 617 knowledge-seeking turns and mixed them with those non-knowledge-seeking turns in the original test set to form the contrast set. 

Table \ref{tab:contrast-set-data-samples} shows several data samples for the newly curated contrast set. These user queries collected from real users are much more diverse than those in the benchmark test set of DSTC9 Track 1 dataset. Among these examples, the user query of ``How much do you charge for parking?'' is actually quite challenging for knowledge-seeking turn detection since this query is very close to one of the available API functions that is responsible for checking whether there is free parking. However, in order to answer this query, we still need to invoke the knowledge module to retrieve external unstructured knowledge. 

\begin{table}[!htpb]
\centering
\small
\begin{tabular}{lccc} 
\toprule
      & Pos   & Neg   & All   \\ \midrule
Train Set & 19,184 & 52,164 & 71,348 \\
Valid Set & 2,673  & 6,990  & 9,663  \\
Test Set & 1,981  & 2,200  & 4,181 \\
Contrast Set  & 617  & 2,200  & 2,817 \\
\bottomrule
\end{tabular}
\caption{Statistics of the knowledge-seeking turn detection benchmark dataset. Pos: knowledge-seeking turns; Neg: non-knowledge-seeking turns.}
\label{tab:data-stats}
\end{table}

\begin{table}[t]
\centering
\small
\begin{tabular}{p{1.2cm}p{5.6cm}}
\toprule
\textbf{Domains}    & \textbf{Examples}                                                               \\ \midrule
Attraction & Is it necessary to buy tickets in advance?                             \\
Attraction & How long it could take to see it all ? 4 hours it would be enough?     \\
Hotel      & Is there a minimum check in age?                                       \\
Hotel      & How much do you charge for parking?                                    \\
Restaurant & Would there be room for a stroller with a sleeping baby during dinner? \\
Restaurant & Can I order crab cakes take out for eight servings ?                  
\\ \bottomrule
\end{tabular}
\caption{Examples of newly collected user questions in the contrast set. These user queries collected from real users are much more diverse than those in the benchmark test set of DSTC9 Track 1 dataset.}
\label{tab:contrast-set-data-samples}
\end{table}

\subsection{Baselines and Settings}

The baselines are 1) the best performing model in the DSTC9 Track 1 competition~\citep{kim2021beyond}, which is a fine-tuned RoBERTa-Large model~\citep{DBLP:journals/corr/abs-1907-11692} on the training set. 
2) Fine-tuned RoBERTa-Large-NLI (obtained by fine-tuning RoBERTa-Large on SNLI and MultiNLI datasets) and DistilBERT-Base-NLI-STSB (obtained by fine-tuning DistilBERT-Base on SNLI, MultiNLI, and STS-B datasets) on the training set.

The sentence encoder $E$ we used is DistilBERT-Base-NLI-STSB~\citep{reimers-2019-sentence-bert}.\footnote{https://github.com/UKPLab/sentence-transformers} The threshold $\eta$ is chosen based on the highest F1 score on the development set. 
For the density estimator, we have tried OC-SVM, KDE with various kinds of kernels, and GMM, and we find GMM performs the best and its inference time is the lowest. We set the number of components to 1 for GMM. Dimensionality $L$ is set as 650 for PCA transformation by tuning on the development set. Details of comparison and tuning results are in the appendix. 
For evaluation metrics, we report precision (P), recall (R), and F1 scores.

\begin{table*}[t]
\centering
\small
\resizebox{0.98\textwidth}{!}{
\begin{tabular}{llcccccccc}
\toprule
\multirow{2}{*}{Learning Schema}                                                                           & \multirow{2}{*}{Sentence Encoder} & \multirow{2}{*}{\begin{tabular}[c]{@{}c@{}}Model\\  size\end{tabular}} & \multirow{2}{*}{\begin{tabular}[c]{@{}c@{}}Trainable \\ Parameters\end{tabular}} & \multicolumn{3}{c}{Test Set (\%)} & \multicolumn{3}{c}{Contrast Set (\%)} \\
                                                                                                           &                                   &                                                                        &                                                                                  & P         & R         & F1        & P           & R          & F1         \\ \midrule
\multirow{3}{*}{Standard Fine-tuning}                                                                      & RoBERTa-Large                 & 355M                                                                   & 355M                                                                             & 99.19     & 92.88     & 95.93     & 96.61       & 69.37      & 80.75      \\
& RoBERTa-Large-NLI                 & 355M                                                                   & 355M                                                                             & 99.46     & 92.28     & 95.73     & 97.54       & 64.18      & 77.42      \\
                                                                                                           & DistilBERT-Base-NLI-STSB          & 66M                                                                    & 66M                                                                              & 98.92     & 92.78     & 95.75     & 95.36       & 66.67      & 78.44      \\ \midrule
\methodname & DistilBERT-Base-NLI-STSB   & 66M                                                                    & 3K      & 97.76     & 94.65     & \textbf{96.18}     & 86.98       & 94.17      & \textbf{90.43}                
\\ \bottomrule
\end{tabular}}
\caption{Performance on the original test set and contrast set when all knowledge-seeking turns data are used for training. Trainable parameters refer to those parameters that are updated for learning knowledge-seeking turns.} 
\label{tab:main-results-full-supervised}
\vspace{-5mm}
\end{table*}

\section{Results \& Discussion}

\subsection{Main Results}

\paragraph{Full supervised setting}
Table \ref{tab:main-results-full-supervised} summarizes the comparison of our method {\methodname} with baselines where all knowledge-seeking turn samples in the DSTC9 Track 1 training set are used for training. 
{\methodname} has two advantages: (1) Once the first step of adaptive pre-training on non-knowledge-seeking turns is done, it only needs to update less than 3K parameters of the density estimator for learning the knowledge-seeking turns, but it can still achieve superior performance on the test set; (2) It can be better generalized to the new contrast set that has distribution shift with respect to the training data.

\paragraph{Low-resource setting}

\begin{figure}[hptb]
\includegraphics[width=2.8in]{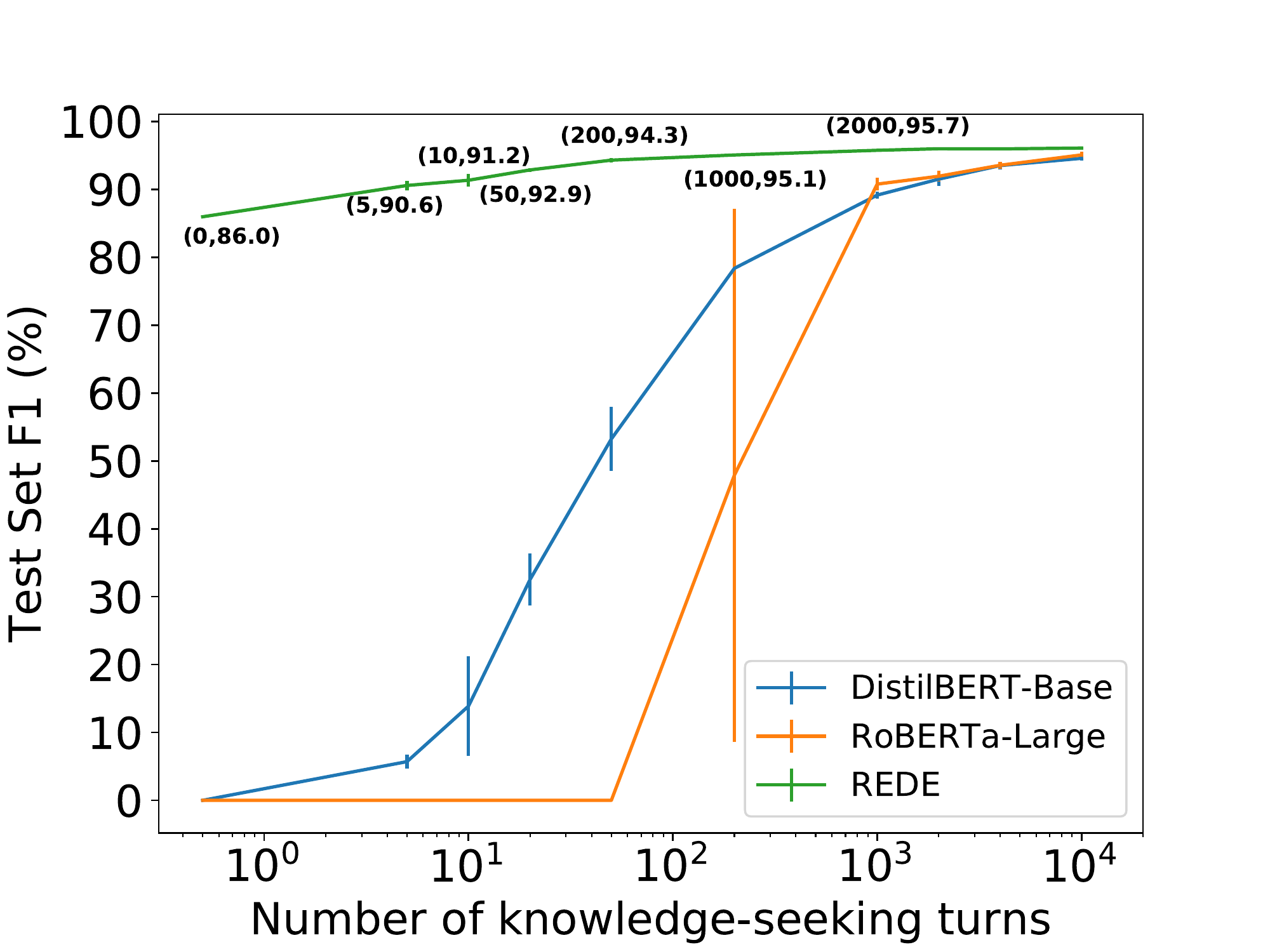}
\caption{F1 score plots with error bar on the test set with different numbers of knowledge-seeking turns used for training Full name of DistilBERT-Base is DistilBERT-Base-NLI-STSB.} 
\label{fig:low-resource-comparison}
\end{figure}

We are more interested in exploring how our method performs under the low-resource setting compared with baselines. Therefore, we sub-sampled different numbers of knowledge-seeking turn samples and kept using all non-knowledge-seeking turn samples. We then trained the model and obtained F1 scores on the test set. We performed five times of random sub-sampling and report the average and standard deviation in Figure~\ref{fig:low-resource-comparison}. Since the error bar of {\methodname} is too small to be seen, we further provide the complete results in Table \ref{tab:full-results-low-resource}. 
As we can see, {\methodname} is always superior than baselines for all sub-sampling ratios. The performance gap is larger when fewer examples are used. Most notably, for the zero-shot setting without using any knowledge-seeking turns, {\methodname} can still achieve 85.95\% of F1 score. 
For comparison, in the zero-shot setting, we also tested Local Outlier Factor (LOF)~\citep{breunig2000lof}, which was used in ~\citep{kim-etal-2020-beyond}, and obtained an F1 score of 73.78\% on the test set, which is much lower than our proposed density estimation method.
Under the few-shots setting such as 5-shots and 10-shots, {\methodname} can obtain more than 90\% of F1, whereas other supervised baselines' scores are under 20\%. 

\begin{table}[h]
\small
\centering
\resizebox{0.98\columnwidth}{!}{
\begin{tabular}{lccc}
\toprule
Samples & DistillBERT & RoBERTa & {\methodname} \\ \midrule
5          &    $5.70\pm0.99$         & $0.00\pm0.00$        & $90.56\pm0.69$                               \\
10         &    $13.88\pm7.37$         &  $0.00\pm0.00$       &     $91.34\pm0.92$                           \\
20         &    $32.51\pm3.84$         &    $0.00\pm0.00$     &     $92.85\pm0.26$              \\
50         &    $53.22\pm4.70$         &    $0.00\pm0.00$     &   $94.29\pm0.35$               \\
200        &    $78.35\pm2.57$          &   $47.87\pm39.28$      &  $95.05\pm0.20$                    \\
1,000       &   $89.14\pm0.49$          &  $90.76\pm0.94$       &   $95.74\pm0.13$              \\
2,000       &   $91.49\pm0.97$          &   $91.92\pm0.79$      & $95.97\pm0.12$                  \\
4,000       &   $93.49\pm0.53$          &   $93.55\pm0.49$      & $95.95\pm0.18$                               \\
10,000      &   $94.56\pm0.32$          &   $95.06\pm0.46$      & $96.07\pm0.17$                 \\ \bottomrule
\end{tabular}}
\caption{Averaged F1 score and standard deviation under the low-resource setting by randomly sub-sampling different number of knowledge-seeking turns for five times. DistillBERT is DistillBERT-Base-NLI-STSB while RoBERTa is RoBERTa-Large.}
\label{tab:full-results-low-resource}
\end{table}

\subsection{Analysis}

\subsubsection{Effect of MLM Adaptation}

As shown in Table \ref{tab:ablation-for-mlm}, after removing the MLM adaptation step, our method has significant performance degradation, especially for the contrast set, indicating the importance of adapting the general pre-trained model to the target dataset via unsupervised learning. We have also tried adopting contrastive learning for such unsupervised adaptation (i.e., SimCSE), which has shown state-of-the-art performance for unsupervised representation learning~\citep{gao2021simcse}. Results in Table \ref{tab:ablation-for-mlm} show that it is worse than MLM. The reason could be that contrastive learning used in SimCSE would lead to more uniform and dispersed distribution over the latent space, however, the density estimation based OOD detection favors more dense and collapsed distribution for in-domain data.

\begin{table}[h]
\centering
\small
\resizebox{0.98\columnwidth}{!}{
\begin{tabular}{lrr}
\toprule
Settings                 & Test Set F1 (\%) & Contrast Set F1 (\%) \\ \midrule
{\methodname}                         & \textbf{96.18}      & \textbf{90.43}          \\
\quad no MLM                    & 93.49      & 75.65          \\
\quad MLM $\rightarrow$ SimCSE & 92.00      & 74.39         \\
\bottomrule
\end{tabular}}
\caption{Ablation study for MLM adaptation by removing it or replacing it with SimCSE (a contrastive learning method). All training samples are used here.}
\label{tab:ablation-for-mlm}
\end{table}

\begin{table}[h]
\centering
\small
\begin{tabular}{lccc} 
\toprule
Dimensions      &  Zero-shot  & Ten-shot   & Full-shot   \\ \midrule
Top 5 & 65.67 & 76.64 & 78.38 \\
Top 50 & 71.04  & 82.23  & 92.40  \\
Top 500 & 77.16 & 91.73 & 96.32 \\
All (768)  & 77.05  & 92.37 & 96.09 \\
\bottomrule
\end{tabular}
\caption{F1 score on test set for top 5, 50, 500, and all principle components under three different settings: zero-shot (PCA over non-knowledge-seeking turns), ten-shot (PCA over ten knowledge-seeking turns), and full-shot (PCA over all knowledge-seeking turns).}
\label{tab:PCA}
\end{table}

\begin{figure}[h]
\centering
     \begin{subfigure}[b]{0.48\columnwidth}
         \centering
         \includegraphics[width=\columnwidth]{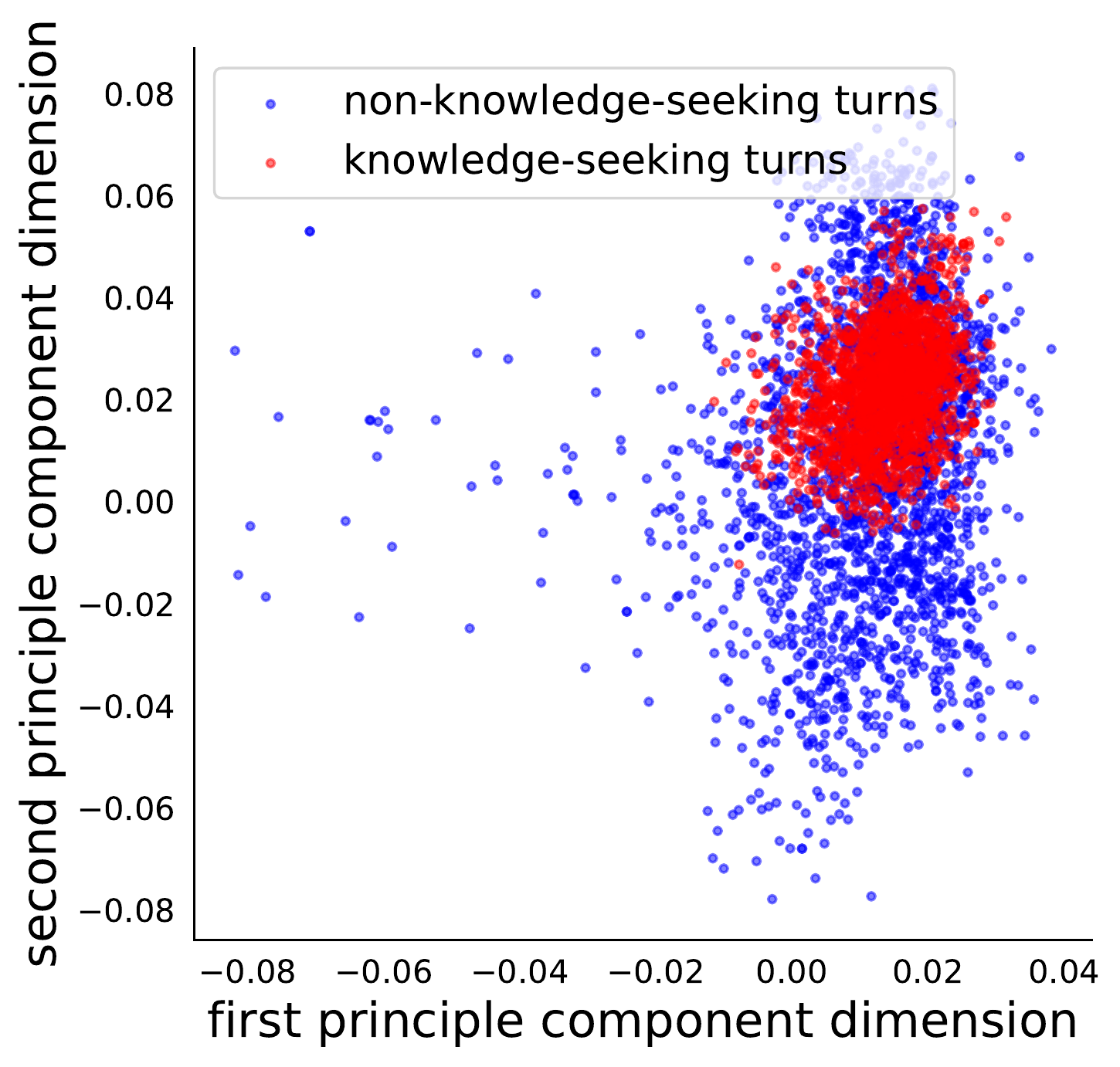}
         \caption{PCA with non-knowledge-seeking turns, F1 = 68.59\%.}
         \label{fig:PCA_without_KS}
     \end{subfigure}
    \begin{subfigure}[b]{0.48\columnwidth}
         \centering
         \includegraphics[width=\columnwidth]{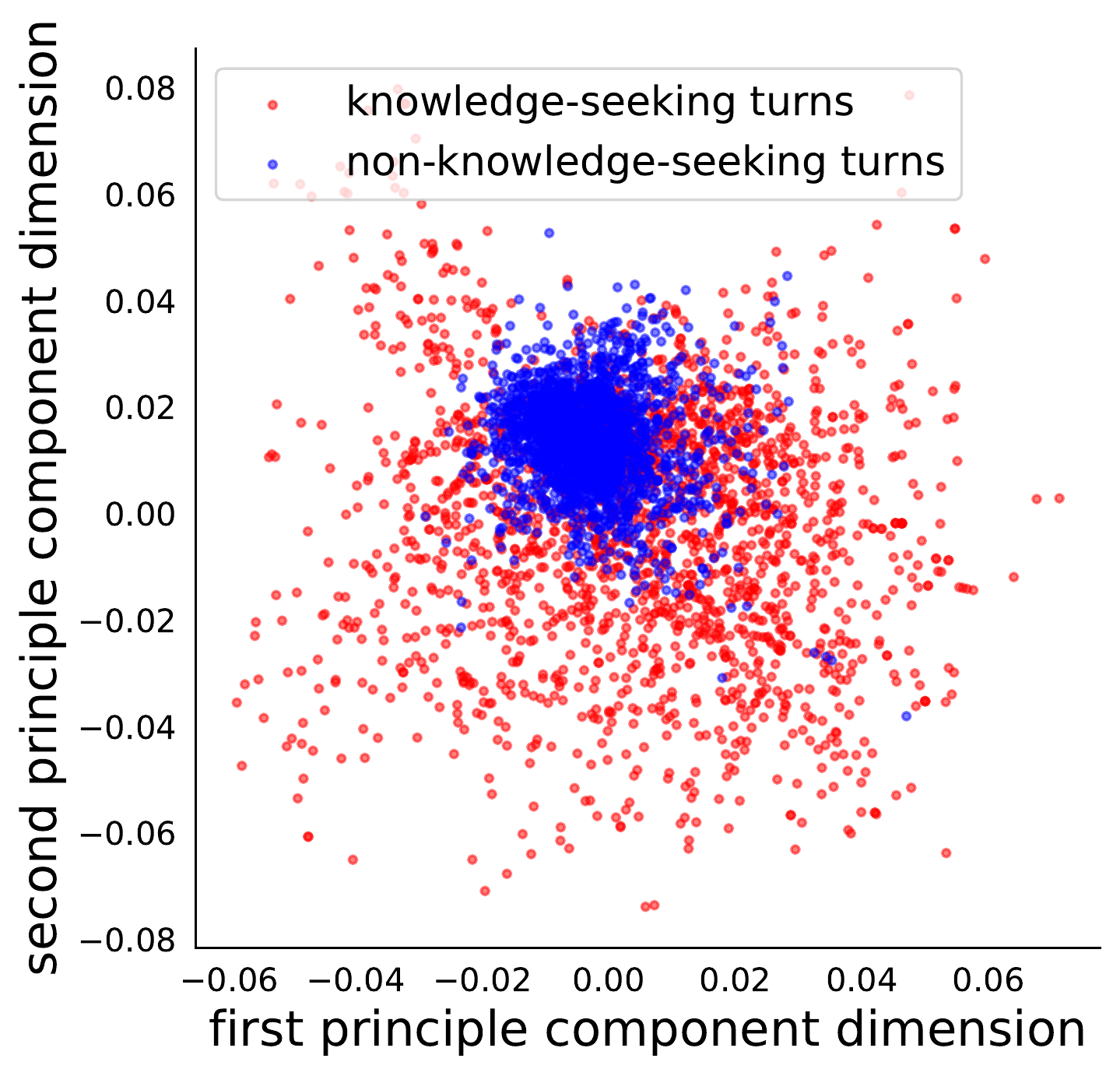}
         \caption{PCA with knowledge-seeking turns, F1 = 78.49\%.}
         \label{fig:PCA_with_KS}
     \end{subfigure}
\caption{Scatter plot using top two principle components of PCA on test samples. F1 score is measured on the test set with top two dimensions only.}
\label{fig:PCA}
\end{figure}

\begin{table*}[t]
\centering
\small
\begin{tabular}{lccccccc}
\toprule
\multirow{2}{*}{Estimator} & \multicolumn{4}{c}{Test Set (\%)}           & \multicolumn{3}{c}{Contrast Set (\%)} \\
                           & P     & R     & F1    & Inference Time (s) & P         & R         & F1       \\ \midrule
OC-SVM                     & 92.30 & 88.34 & 90.28 & 74.36         & 68.81     & 88.33     & 77.36    \\
KDE-Gaussian               & 92.81 & 91.17 & 91.98 & 377.43         & 72.49     & 88.82     & 79.83    \\
KDE-Exponential            & 92.29 & 91.87 & 92.08 & 373.76         & 73.18     & 88.01     & 79.91    \\
GMM                        & \textbf{97.76}     & \textbf{94.65}     & \textbf{96.18}  & \textbf{0.07}    & \textbf{86.98}       & \textbf{94.17}      & \textbf{90.43}                 
\\ \bottomrule
\end{tabular}
\caption{Comparison of different density estimators. Inference time is measure on the whole test set using the same machine. We have also tried other kernels for the KDE estimator, such as `tophat', `epanechnikov', `linear', and `cosine', but they all perform poorly.}
\label{tab:comparison-estimators}
\end{table*}

\noindent
\subsubsection{Understanding PCA Transformation}
\label{sec:PCA-transformation}

In Section~\ref{sec:rt}, the sentence representation is transformed with PCA learned from knowledge-seeking turns. 
Table~\ref{tab:PCA} shows the F1 score on the test set using top $L$ principle components with PCA learned using different data.  
Overall, we can see that PCA with knowledge-seeking turns achieves better performance, and using more principle components is always beneficial.  
PCA is well-known to help construct new subspaces by maximizing the global variance. Intuitively,  by learning PCA over knowledge-seeking turns,  we expect the manifolds on knowledge-seeking turns to spread out and non-knowledge seeking turns condense.  
Figure~\ref{fig:PCA} shows the scatter plot of the top two principle components of transformed features. 
In Figure~\ref{fig:PCA_without_KS}, we learn PCA from non-knowledge-seeking turns, which results in the manifold of knowledge-seeking turns (red dots) to be within that of non-knowledge-seeking turns (blue dots). It hurts the performance since the density estimation is performed over non-knowledge-seeking turns, as confirmed by the zero shot result in Table~\ref{tab:PCA} in comparison to that in Fig~\ref{fig:low-resource-comparison}.
In contrast, in Figure~\ref{fig:PCA_with_KS}, we learn PCA with knowledge-seeking turns, which makes knowledge-seeking turns (red dots) spread out and non-knowledge-seeking turns (blue dots) condense. 
By estimating the density of this condensed blue area, we obtain higher F1 score because all the red dots falling outside of the region of blue dots will be classified as out-of-distribution correctly. 

\begin{table}[t]
\centering
\small
\resizebox{0.98\columnwidth}{!}{
\begin{tabular}{lccc}
\toprule
Components \# & Dev Set F1 & Test Set F1 & Contrast Set F1 \\ \midrule
1                & 98.71 & 96.18            & 90.43                \\
2                & 98.88 & 95.82            & 90.61                \\
3                & 98.97 & 96.12            & 90.35                \\
4                & 99.03 &  96.04            & 89.72               
\\ \bottomrule
\end{tabular}}
\caption{Comparison of performance (in percentage) by using different number of components for the GMM estimator.}
\label{tab:gmm-number-components}
\end{table}

\begin{figure}[h]
\includegraphics[width=\columnwidth]{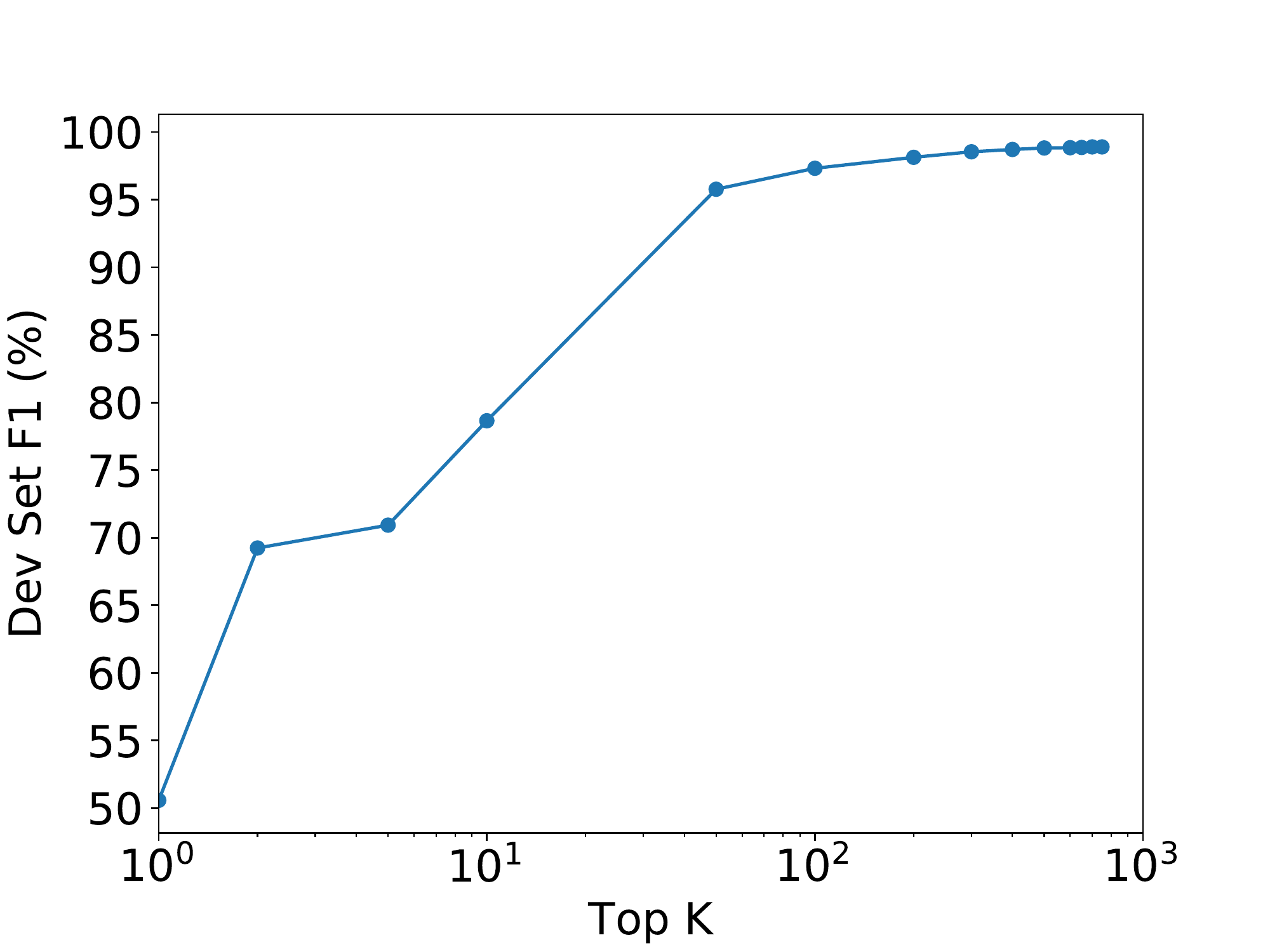}
\caption{Development set F1 scores by retraining different values of first $L$ columns of $W$. Full dimension is 768.}
\label{fig:different-K-values}
\end{figure}

\subsubsection{Comparison of Density Estimators}

For the density estimator, we have tried OC-SVM, KDE with various kinds of kernels, and GMM, which are summarized in Table \ref{tab:comparison-estimators}. All these estimators are implemented using Scikit-Learn library.\footnote{https://scikit-learn.org/stable/} From Table \ref{tab:comparison-estimators}, we see that GMM performs the best while being the fastest for inference, therefore we chose it as the density estimator in our work.

Table \ref{tab:gmm-number-components} shows the performance under different number of components for the GMM density estimator. From it, we see that the number of components has minor influence on the performance so we decide to use 1 as the number of components in this work.

\subsubsection{Effects of $L$}

We can retrain only the first $L$ columns of $W$ for the PCA transformation, which can help us reduce the dimension of transformed representation vector $\tilde{e}$. Figure \ref{fig:different-K-values} shows the development set performance under different values of $L$ when all knowledge-seeking turns are used for training. We see that the first 50 dimensions can achieve over 95\% F1 score and 300 dimensions are already enough to realize the peak performance, whereas the full dimension is 768. 


\vspace{-0.0in}
\section{Conclusion}
\vspace{-0.08in}
In this work, we propose a novel method {\methodname} based on domain-adapted representation learning and density estimation for knowledge-seeking turn detection in tasked-orientated dialogue systems. Compared with previous SOTA models, {\methodname} can achieve comparable performance in the full supervised setting and significantly superior performance for the low-resource setting. Besides, {\methodname} has much better generalization capability onto a new contrast set we curated.
\bibliography{anthology,custom}
\bibliographystyle{acl_natbib}








\end{document}